\documentclass{article} % For LaTeX2e
\usepackage{iclr2026_conference,times}
\usepackage{multirow}    
\usepackage{booktabs}    
\usepackage{makecell}     
\usepackage{array}       
\usepackage{xcolor}       
\usepackage{graphicx}     

\usepackage{amsmath,amsfonts,bm}

\def\eqref#1{equation~\ref{#1}}
\def\1{\bm{1}}

\DeclareMathAlphabet{\mathsfit}{\encodingdefault}{\sfdefault}{m}{sl}
\SetMathAlphabet{\mathsfit}{bold}{\encodingdefault}{\sfdefault}{bx}{n}

\usepackage{hyperref}
\usepackage{url}
\usepackage{subcaption}

\title{SegDINO: An Efficient Design for Medical and Natural Image Segmentation with DINO-V3}

\author{
Sicheng Yang$^{1}$, Hongqiu Wang$^{1}$, Zhaohu Xing$^1$, 
Sixiang Chen$^1$,
Lei Zhu$^{1,2\thanks{Lei Zhu (leizhu@ust.hk) is the corresponding author.}}$ \\
$^1$The Hong Kong University of Science and Technology (Guangzhou), China \\
$^2$The Hong Kong University of Science and Technology, China
}

\iclrfinalcopy % Uncomment for camera-ready version, but NOT for submission.
\begin{document}

\maketitle
\begin{abstract}
The DINO family of self-supervised vision models has shown remarkable transferability, yet effectively adapting their representations for segmentation remains challenging. Existing approaches often rely on heavy decoders with multi-scale fusion or complex upsampling, which introduce substantial parameter overhead and computational cost. In this work, we propose \textbf{SegDINO}, an efficient segmentation framework that couples a frozen DINOv3 backbone with a lightweight decoder. SegDINO extracts multi-level features from the pretrained encoder, aligns them to a common resolution and channel width, and utilizes a lightweight MLP head to directly predict segmentation masks. This design minimizes trainable parameters while preserving the representational power of foundation features. Extensive experiments across six benchmarks, including three medical datasets (TN3K, Kvasir-SEG, ISIC) and three natural image datasets (MSD, VMD-D, ViSha), demonstrate that SegDINO consistently achieves state-of-the-art performance compared to existing methods. Code is available at \url{https://github.com/script-Yang/SegDINO}.
\end{abstract}

\section{introduction}
Image segmentation plays a central role in image analysis, serving as the foundation for downstream tasks such as object recognition~\citep{minaee2021image}, scene understanding~\citep{jain2023oneformer,wang2024video}, and computer-aided diagnosis~\citep{azad2024medical,wang2024dual,wang2025serp}. Despite remarkable progress achieved by convolutional networks~\citep{long2015fully,ronneberger2015u}, 
transformer-based models~\citep{strudel2021segmenter,li2024transformer}, 
diffusion-based architectures~\citep{amit2021segdiff,wu2024medsegdiff}, 
and Mamba-based frameworks~\citep{ma2024u,xing2024segmamba}, 
these approaches often struggle to achieve strong generalization when training data are limited~\citep{zhang2021understanding}. Recent SAM-based segmentation models~\citep{kirillov2023segment,mazurowski2023segment} offer powerful zero-shot capabilities but typically require extensive fine-tuning for downstream tasks, leading to inefficiency~\citep{zhang2024efficientvit}. Moreover, even in the frozen setting, the computational overhead of SAM models is substantial, making them less suitable for lightweight or resource-constrained applications~\citep{zhao2023fast}. Consequently, the design of high-performance yet efficient segmentation frameworks remains an open challenge.

With the emergence of self-supervised foundation models~\citep{caron2021emerging,he2022masked}, pretrained vision backbones have become increasingly prevalent for dense prediction. Rather than training encoders from scratch, recent segmentation methods leverage large-scale pretrained representations to capture rich semantics and structural priors~\citep{zhou2024image}. 
% This paradigm substantially improves generalization and reduces reliance on labeled data.
In contrast to SAM-based models, self-supervised vision models achieve a favorable balance by maintaining relatively modest parameter counts while extracting high-quality semantic features~\citep{simeoni2025dinov3}, making them especially appealing for segmentation tasks.

Among various self-supervised foundation models, the DINO family has demonstrated exceptional transferability across a wide range of visual tasks~\citep{wang2025dinov3,gao2025dinoUNet}. DINO~\citep{caron2021emerging} and DINOv2~\citep{oquab2023dinov2} have been widely adopted for representation learning~\citep{zhu2024scaling, wang2025vggt}, providing robust multi-scale features suitable for detection~\citep{damm2025anomalydino} and segmentation~\citep{ayzenberg2024dinov2}. Most recently, DINOv3~\citep{simeoni2025dinov3} introduced significant improvements in pretraining strategies and architectural refinements, achieving stronger invariance and scalability, and establishing itself as a state-of-the-art pretrained backbone.

However, effectively adapting DINO-based representations for segmentation tasks still remains a non-trivial challenge. Existing methods often employ relatively heavy decoders, such as multi-scale fusion modules~\citep{gao2025dinoUNet} or complex upsampling pipelines~\citep{yang2025unimatch}, which introduce substantial parameter overhead and computational cost. Such decoder complexity consequently offsets the efficiency advantages of frozen pretrained encoders and poses obstacles to deployment in resource-constrained settings~\citep{xie2021segformer}.

To address these limitations, we propose SegDINO, a segmentation framework that couples a frozen DINOv3 backbone with a lightweight decoder. SegDINO leverages the DINO backbone to extract semantically rich features and employs a light MLP-based head to directly predict segmentation masks. This design minimizes the trainable parameter burden while preserving representational power from the foundation encoder. Extensive experiments on both medical  and natural image segmentation benchmarks demonstrate that SegDINO achieves competitive or superior accuracy compared to baselines while offering significant efficiency advantages.

% \section{Related Work}
\section{Methodology}
\subsection{Overview}
As illustrated in Fig.~\ref{fig:overview}, an input image is fed to a pretrained, frozen DINOv3 model to extract multi-layer features. The selected features are lightly upsampled to a common spatial resolution, concatenated along the channel dimension, and passed to a lightweight decoder to produce the final segmentation mask. During training, only the decoder is updated.

\subsection{Encoder backbone}
We adopt a pretrained DINOv3 Vision Transformer~\citep{simeoni2025dinov3} as the encoder and freeze all its parameters throughout training. 
Given an input image $\mathbf{x}\in\mathbb{R}^{H\times W\times 3}$ and a patch size $p$, the encoder divides $\mathbf{x}$ into $N=(H/p)\times(W/p)$ patches, each of which is linearly projected into a $d$-dimensional token representation. 
The resulting patch-token matrix is denoted as $\mathbf{Z}^{(0)}\in\mathbb{R}^{N\times d}$. 
Following the DINOv3 design, the backbone is a ViT with $L$ Transformer blocks. Let $\mathcal{B}_\ell$ denote the $\ell$-th Transformer block; the token sequence is updated as
\begin{align}
\mathbf{Z}^{(\ell)}=\mathcal{B}_\ell\! \ \big(\mathbf{Z}^{(\ell-1)}\big),\qquad \ell=1,\dots,L.
\end{align}
To harvest both low-level structure and high-level semantics, we collect intermediate token matrices from a subset of layers
\begin{align}
\mathcal{L}=\{\ell_1,\ell_2,\dots,\ell_K\}\subseteq\{1,\dots,L\}.
\end{align}
For each $\ell_k\in\mathcal{L}$, we directly take the patch tokens $\mathbf{Z}_{\mathrm{p}}^{(\ell_k)}\in\mathbb{R}^{N\times d}$ from the ViT output and discard any non-patch tokens (e.g., class or register tokens). 
The encoder output is the multi-level token set
\begin{align}
\mathcal{F}=\big\{\mathbf{Z}_{\mathrm{p}}^{(\ell_k)}\big\}_{k=1}^{K}.
\end{align}
These patch-token features are forwarded to the lightweight decoder (see Fig.~\ref{fig:overview}) to produce segmentation representations. 
Freezing encoder stabilizes training and yields transferable features while keeping the trainable burden on the light decoder head.

\subsection{L-Decoder}
The proposed \textbf{L}ight-Decoder follows a reform strategy similar to the upsampling and channel integration design in ~\citep{ranftl2021vision}, where multi-level features are progressively aligned to a common spatial resolution and channel width. Let $\widetilde{\mathbf{Z}}^{(\ell_k)}\in\mathbb{R}^{N\times C}$ denote the reformulated feature map obtained from each selected layer $\ell_k\in\mathcal{L}$. These features are concatenated along the channel dimension to form
\begin{align}
\mathcal{H}=\mathrm{Concat}\big(\widetilde{\mathbf{Z}}^{(\ell_1)},\widetilde{\mathbf{Z}}^{(\ell_2)},\dots,\widetilde{\mathbf{Z}}^{(\ell_K)}\big)\in\mathbb{R}^{N\times KC}.
\end{align}
The fused representation $\mathcal{H}$ is then passed through a lightweight decoder $\mathrm{D}_{\theta_d}$, implemented as a multi-layer perceptron (MLP), to produce the final segmentation mask
\begin{align}
\widehat{\mathbf{y}}=\mathrm{D}_{\theta_d}(\mathcal{H}),\qquad \widehat{\mathbf{y}}\in\mathbb{R}^{N\times n_{\text{class}}},
\end{align}
where $n_{\text{class}}$ denotes the number of semantic classes. This lightweight design ensures efficient training while retaining strong representational capacity for dense prediction.
\begin{figure*}[t]
\centering
\includegraphics[width=0.99\linewidth]{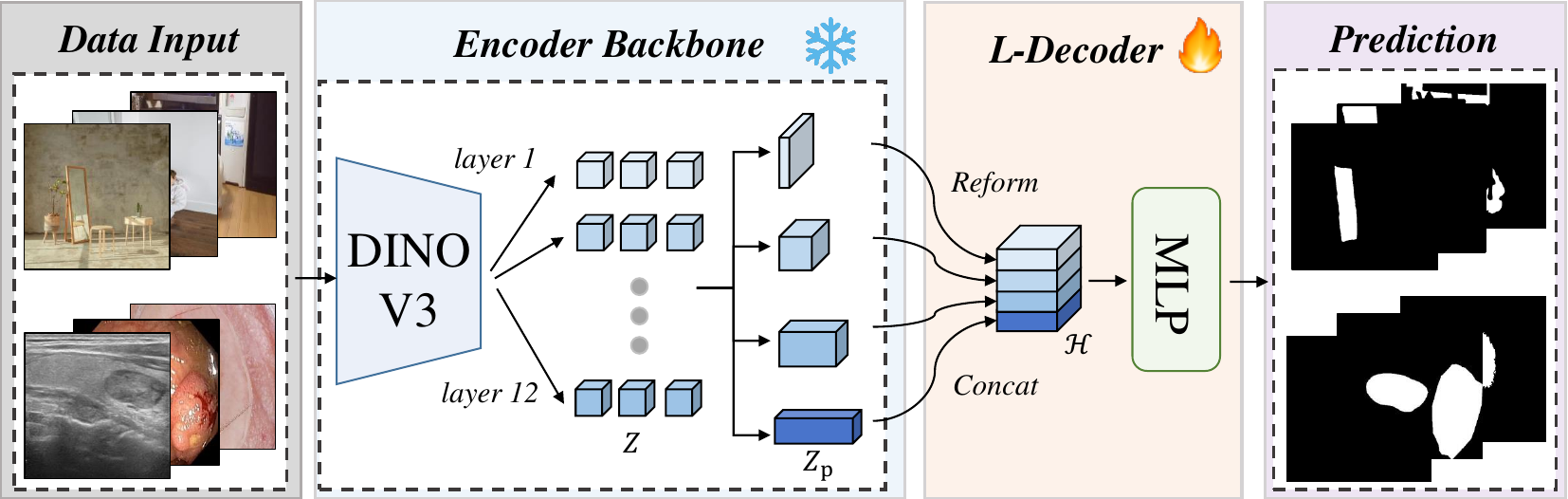}
\caption{SegDINO couples a frozen DINOv3 with a lightweight decoder for efficient segmentation. Multi-layer features from different depths are upsampled, aligned, and concatenated, then passed to a super light MLP head to produce the final segmentation mask.}
\label{fig:overview}
\end{figure*}

\section{experiments}

\subsection{Datasets}

\paragraph{Medical Image Datasets.}
We evaluate our method on three medical image segmentation benchmarks.  
TN3K~\citep{gong2023thyroid} is a large-scale thyroid nodule segmentation dataset, containing 3,493 ultrasound images with pixel-level annotations collected from multiple hospitals.  
Kvasir-SEG~\citep{jha2019kvasir} is a polyp segmentation dataset derived from colonoscopy examinations, consisting of 1,000 images with high-quality expert annotations.  
ISIC~\citep{codella2018skin} is a skin lesion segmentation benchmark, providing 2,750 dermoscopic images annotated for lesion boundaries and covering a wide range of lesion types and acquisition conditions. 

\paragraph{Natural Image Datasets.}
We conduct experiments on three representative benchmarks covering both mirror and shadow segmentation tasks. 
MSD~\citep{yang2019my} is a static image-based mirror segmentation dataset, containing 4,018 annotated images from diverse scenarios such as indoor mirrors, shop windows, and vehicle mirrors. 
VMD-D~\citep{lin2023learning} is the first large-scale video mirror detection dataset, which consists of 269 videos (14,988 frames) with high-resolution annotations, capturing challenging dynamic conditions including camera motion, illumination variations, and multiple mirrors. 
VISHA~\citep{chen2021triple} is a widely used benchmark for video shadow detection, providing 11,685 video frames with fine-grained annotations.

\subsection{Implementation Details}
\paragraph{Experimental Settings.} 
For each dataset, we follow the official training–testing split provided by the organizers to ensure fair comparison. 
All images are resized to $256 \times 256$ for consistent input resolution across models, and normalized using the same mean and standard deviation parameters as in DINOv3~\citep{simeoni2025dinov3}.

We implement all experiments with the PyTorch framework~\citep{paszke2019pytorch}. 
The models are optimized using AdamW~\citep{loshchilov2017decoupled} with a learning rate of $1\times10^{-4}$ and a weight decay of $1\times10^{-4}$. 
Cross-entropy loss is employed as the training objective. 
Training is conducted for 50 epochs with a batch size of 4. 
For SegDINO, the DINO backbone is frozen, and only the decoder parameters are updated. 
In this work, we exclusively adopt the DINOv3-S backbone, from which intermediate features of the 3rd, 6th, 9th, and 12th Transformer layers are extracted. 
All experiments are run on a cloud platform equipped with four NVIDIA RTX A6000 GPUs.

\paragraph{Evaluation Metrics.} For medical image datasets, we employ Dice similarity coefficient (DSC) and IoU to measure overlap between predictions and ground truth, together with the 95th percentile Hausdorff Distance (HD95) to evaluate boundary localization accuracy. For natural image datasets, we adopt intersection over union (IoU), pixel accuracy (Accuracy), F-measure ($F_\beta$)~\citep{lin2023learning}, mean absolute error (MAE), and balanced error rate (BER) to evaluate our method. 
For shadow segmentation, we additionally report Shadow-BER (S-BER) and Non-shadow-BER (N-BER) for class-specific assessment~\citep{vicente2017leave}.

\begin{table*}[htbp]
  \centering
  \caption{Comparison with state-of-the-art models for medical image segmentation.}
  \setlength{\tabcolsep}{1.8mm}
  \small
  \begin{tabular}{lccccccccc}
    \toprule
    \multirow{2}[4]{*}{Methods} 
      & \multicolumn{3}{c}{TN3K}    
      & \multicolumn{3}{c}{Kvasir-SEG}    
      & \multicolumn{3}{c}{ISIC} \\
\cmidrule{2-10}
      & DSC$\uparrow$ & IoU$\uparrow$ & HD95$\downarrow$
      & DSC$\uparrow$ & IoU$\uparrow$ & HD95$\downarrow$
      & DSC$\uparrow$ & IoU$\uparrow$ & HD95$\downarrow$ \\
    \midrule
    U-Net        & 0.7945 & 0.7065 & 24.59 & 0.7916 & 0.7029 & 41.58 & 0.8187 & 0.7295 & 25.12 \\
    SegNet             & 0.7924 & 0.7001 & 22.74 & 0.8415 & 0.7565 & 25.89 & 0.8327 & 0.7446 & 21.41 \\
    R2U-Net     & 0.6886 & 0.5935 & 27.89 & 0.7367 & 0.6328 & 45.64 & 0.8102 & 0.7134 & 25.36 \\
    Att-UNet    & 0.8015 & 0.7116 & 24.64 & 0.8016 & 0.7202 & 31.92 & 0.8275 & 0.7372 & 26.12 \\
    TransUNet          & 0.8027 & 0.7081 & 23.95 & 0.8054 & 0.7093 & 37.67 & 0.8186 & 0.7230 & 24.76 \\
    U-NeXt      & 0.7285 & 0.6245 & 31.40 & 0.6271 & 0.5164 & 58.32 & 0.8230 & 0.7327 & 23.63 \\
    U-KAN       & 0.7866 & 0.6960 & 24.49 & 0.7217 & 0.6381 & 36.92 & 0.8341 & 0.7462 & 23.57 \\
    \midrule
    \textbf{SegDINO}   & \textbf{0.8318} & \textbf{0.7443} & \textbf{18.62} & \textbf{0.8765} & \textbf{0.8064} & \textbf{20.80} & \textbf{0.8576} & \textbf{0.7760} & \textbf{17.80} \\
    % \textbf{SegDINO-B}   & \textbf{0.8351} & \textbf{0.7464} & \textbf{18.16} & 0.8637 & 0.7903 & 22.75 & 0.8570 & 0.7731 & 17.93 \\
    \bottomrule
  \end{tabular}
  \label{tab:cmp_sota}
\end{table*}

\begin{table}[htbp]
  \centering
  \begin{minipage}{0.47\linewidth}
    \centering
    \caption{Results on the MSD dataset.}
    \label{tab:msd} 
    \setlength{\tabcolsep}{1.8 mm}
    \small
    \begin{tabular}{l c c c c}
      \toprule
      Methods & IoU$\uparrow$ & Acc.$\uparrow$ & $F_\beta\uparrow$ & MAE$\downarrow$ \\
      \midrule
      SegFormer       & 0.879 & 0.953 & 0.915 & 0.038 \\
      Mask2Former     & 0.883 & 0.958 & 0.917 & 0.036 \\
      MirrorNet       & 0.845 & 0.948 & 0.892 & 0.044 \\
      PMDNet          & 0.847 & 0.952 & 0.898 & 0.033 \\
      VCNet           & 0.884 & 0.958 & 0.917 & 0.029 \\
      SANet           & 0.871 & 0.951 & 0.914 & 0.032 \\
      HetNet          & 0.888 & 0.964 & 0.918 & 0.030 \\
      CSFwinformer    & 0.875 & 0.960 & 0.918 & 0.032 \\
      \midrule
    \textbf{SegDINO} & \textbf{0.942} & \textbf{0.985} & \textbf{0.971} & \textbf{0.015} \\

      \bottomrule
    \end{tabular}
  \end{minipage}%
  \hfill
  \begin{minipage}{0.47\linewidth}
    \centering
    \caption{Results on the VMD-D dataset.}
    \label{tab:vmd} 
    \setlength{\tabcolsep}{1.8 mm}
    \small
    \begin{tabular}{l c c c c}
      \toprule
      Methods & IoU$\uparrow$ & Acc.$\uparrow$ & $F_\beta\uparrow$ & MAE$\downarrow$ \\
      \midrule
      TVSD            & 0.480 & 0.875 & 0.746 & 0.125 \\
      STICT           & 0.164 & 0.809 & 0.530 & 0.198 \\
      Sc-Cor   & 0.512 & 0.863 & 0.696 & 0.137 \\
      Scotch-Soda & 0.587 & 0.870 & 0.706 & 0.124 \\
      HFAN            & 0.459 & 0.876 & 0.706 & 0.121 \\
      STCN            & 0.445 & 0.859 & 0.670 & 0.140 \\
      GlassNet        & 0.552 & 0.864 & 0.718 & 0.137 \\
      MirrorNet       & 0.580 & 0.864 & 0.724 & 0.135 \\
      PMDNet          & 0.532 & 0.872 & 0.749 & 0.128 \\
      VCNet           & 0.539 & 0.877 & 0.749 & 0.123 \\
      HetNet          & 0.531 & 0.877 & 0.745 & 0.123 \\
      VMD-Net  & 0.567 & 0.895 & 0.787 & 0.105 \\
      \midrule
    \textbf{SegDINO} & \textbf{0.762} & \textbf{0.933} & \textbf{0.852} & \textbf{0.072} \\
      \bottomrule
    \end{tabular}
  \end{minipage}%
\end{table}

\begin{table}[t]
\centering
\caption{Quantitative comparison on the ViSha dataset.}
\label{tab:visha} 
\setlength{\tabcolsep}{1.8mm}
\small
\begin{tabular}{l c c c c c c}
\toprule
Methods  & IoU $\uparrow$ & $F_\beta$ $\uparrow$ & MAE $\downarrow$ & BER $\downarrow$ & S-BER $\downarrow$  & N-BER $\downarrow$ \\
\midrule

STM             & 0.408 & 0.598 & 0.069 & 25.69 & 47.44 & 3.95 \\
COS-Net         & 0.515 & 0.706 & 0.040 & 20.51 & 39.22 & 1.79 \\
MTMT            & 0.517 & 0.729 & 0.043 & 20.29 & 38.71 & 1.86 \\
FSD             & 0.486 & 0.671 & 0.057 & 20.57 & 38.06 & 3.06 \\
TVSD            & 0.556 & 0.757 & 0.033 & 17.70 & 33.97 & 1.45 \\
STICT           & 0.545 & 0.702 & 0.046 & 16.60 & 29.58 & 3.59 \\
Sc-Cor          & 0.615 & 0.762 & 0.042 & 13.61 & 24.31 & 2.91 \\
Scotch-Soda     & 0.640 & 0.793 & 0.029 & 9.06  & 16.26 & 1.44 \\
SATNet          & 0.521 & 0.730 & 0.046 & 21.18 & 38.64 & 3.02 \\
CSFwinformer    & 0.525 & 0.733 & 0.040 & 20.01 & 36.54 & 1.99 \\
VGSD-Net        & 0.548 & 0.733 & 0.052 & 19.98 & 35.24 & 3.67 \\
TBG-Diff & 0.667 & 0.797 & 0.023 & 8.58 & 16.00 & 1.15 \\
\midrule
\textbf{SegDINO}  & \textbf{0.675} & \textbf{0.821} & \textbf{0.017} & \textbf{8.05} & \textbf{14.90} & \textbf{0.82} \\
\bottomrule
\end{tabular}
\end{table}

\subsection{Comparison with Existing Methods}

\paragraph{Comparison on medical image benchmarks.}
We compare SegDINO with a diverse set of state-of-the-art segmentation models, including U-Net~\citep{ronneberger2015u}, SegNet~\citep{badrinarayanan2017segnet}, R2U-Net~\citep{alom2018recurrent}, Attention U-Net~\citep{oktay2018attention}, TransUNet~\citep{chen2021transunet}, U-NeXt~\citep{valanarasu2022unext}, and U-KAN~\citep{li2025u}.

As shown in Table~\ref{tab:cmp_sota}, both variants achieve consistent improvements across TN3K, Kvasir-SEG, and ISIC datasets. On TN3K, SegDINO yields the best Dice score of 0.8318, surpassing the strongest competitor TransUNet by +3\% in DSC, +3.6\% in IoU, and reducing HD95 from 23.95 to 18.62. On Kvasir-SEG, SegDINO achieves the highest performance with a Dice score of 0.8765 and IoU of 0.8064, outperforming the second-best SegNet by +3.5\% in DSC and +5.0\% in IoU, while decreasing HD95 from 25.89 to 20.80. On ISIC, SegDINO again leads with a Dice score of 0.8576 and IoU of 0.7760, improving over the best baseline U-KAN by +2.3\% in DSC and +3.0\% in IoU, and lowering HD95 from 23.57 to 17.80.

\paragraph{Comparison on natural image benchmarks.}  
We conduct comprehensive comparisons on three representative natural image segmentation benchmarks, including MSD for static mirror segmentation, VMD-D for dynamic video mirror detection, and ViSha for video shadow detection. On MSD, our SegDINO is compared with SegFormer~\citep{xie2021segformer}, Mask2Former~\citep{chen2021simple}, MirrorNet~\citep{yang2019my}, PMDNet~\citep{lin2020progressive}, VCNet~\citep{tan2022mirror}, SANet~\citep{guan2022learning}, HetNet~\citep{he2023efficient}, and CSFwinformer~\citep{xie2024csfwinformer}. On VMD-D, we evaluate against TVSD~\citep{chen2021triple}, STICT~\citep{lu2022video}, Sc-Cor~\citep{ding2022learning}, Scotch-Soda~\citep{liu2023scotch}, HFAN~\citep{pei2022hierarchical}, STCN~\citep{cheng2021rethinking}, GlassNet~\citep{lin2021rich}, MirrorNet~\citep{yang2019my}, PMDNet~\citep{lin2020progressive}, VCNet~\citep{tan2022mirror}, HetNet~\citep{he2023efficient}, and VMD-Net~\citep{lin2023learning}. On ViSha, we benchmark against STM~\citep{oh2019video}, COS-Net~\citep{lu2019see}, MTMT~\citep{chen2020multi}, FSD~\citep{hu2021revisiting}, TVSD~\citep{chen2021triple}, STICT~\citep{lu2022video}, Sc-Cor~\citep{ding2022learning}, Scotch-Soda~\citep{liu2023scotch}, SATNet~\citep{huang2023symmetry}, CSFwinformer~\citep{xie2024csfwinformer}, VGSD-Net~\citep{liu2024multi}, and TBGDiff~\citep{zhou2024timeline}.  

Across all three datasets, SegDINO consistently achieves the best results and surpasses existing approaches by clear margins. On MSD (Tab.~\ref{tab:msd}), it outperforms the second-best method HetNet by over 5\% in IoU, over 2\% in accuracy, over 5\% in $F_\beta$. On VMD-D (Tab.~\ref{tab:vmd}), it surpasses the strongest competitor VMD-Net with relative gains of more than 19\% in IoU, over 3\% in accuracy, over 6\% in $F_\beta$. On ViSha (Tab.~\ref{tab:visha}), SegDINO improves over the second-best method TBG-Diff by nearly 1\% in IoU, more than 2\% in $F_\beta$, while also achieving significantly lower BER.  

\begin{figure}[t]
\centering
\begin{subfigure}{0.32\linewidth}
    \centering
    \includegraphics[width=\linewidth]{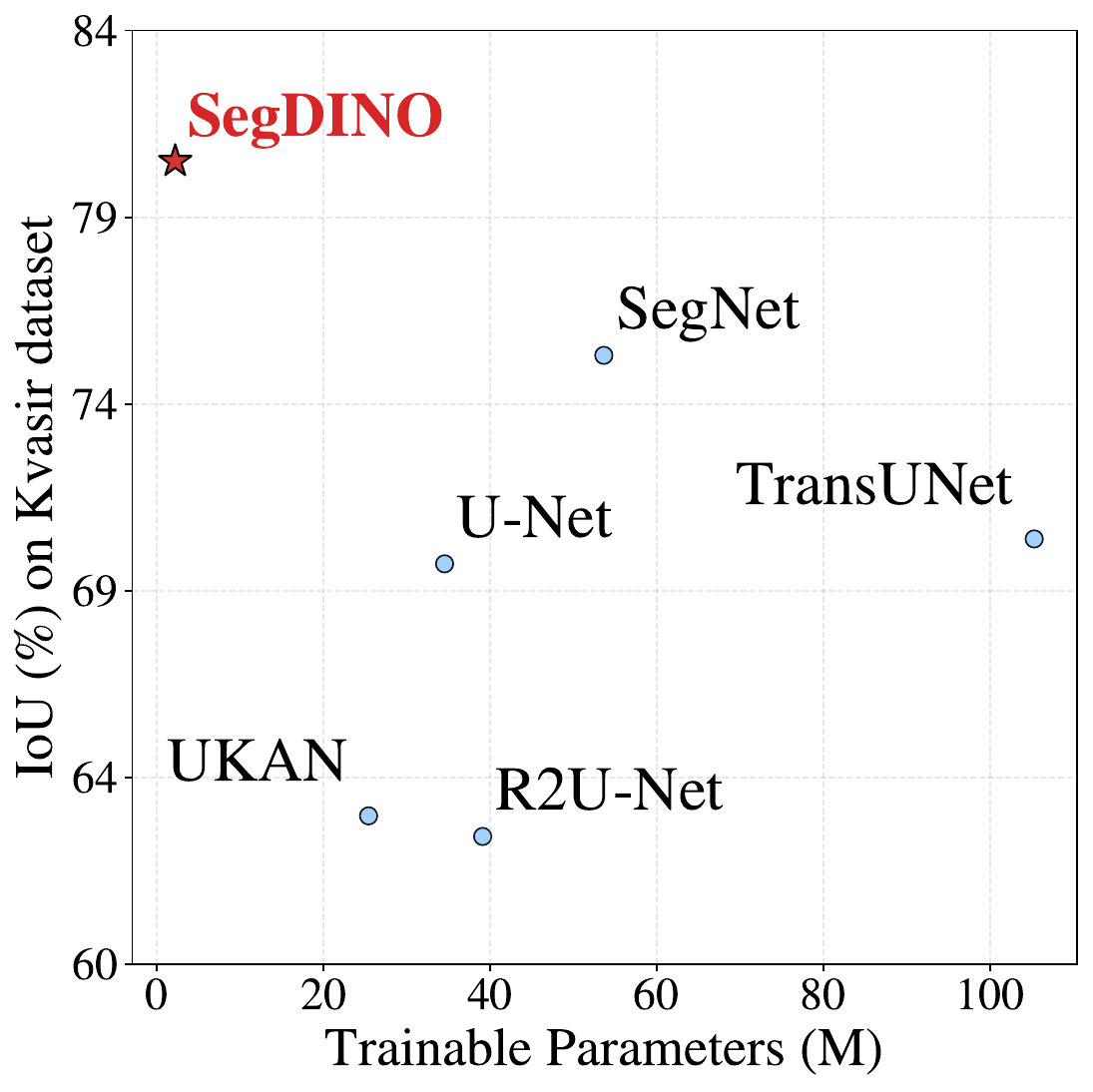}
    \label{fig:overview-medical}
\end{subfigure}
% \hfill
\begin{subfigure}{0.32\linewidth}
    \centering
    \includegraphics[width=\linewidth]{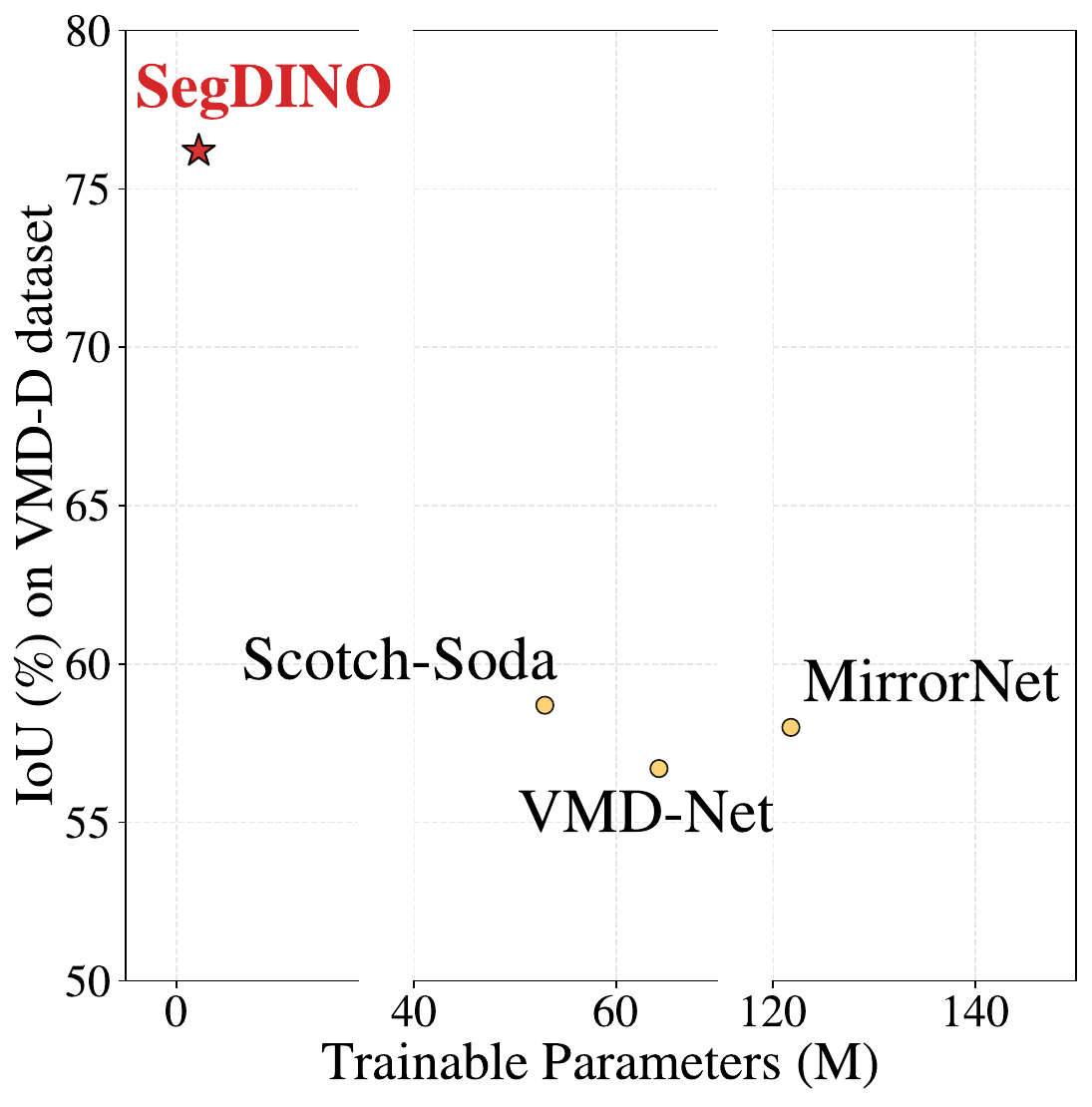}
    \label{fig:overview-natural}
\end{subfigure}
% \hfill
\begin{subfigure}{0.32\linewidth}
    \centering
    \includegraphics[width=\linewidth]{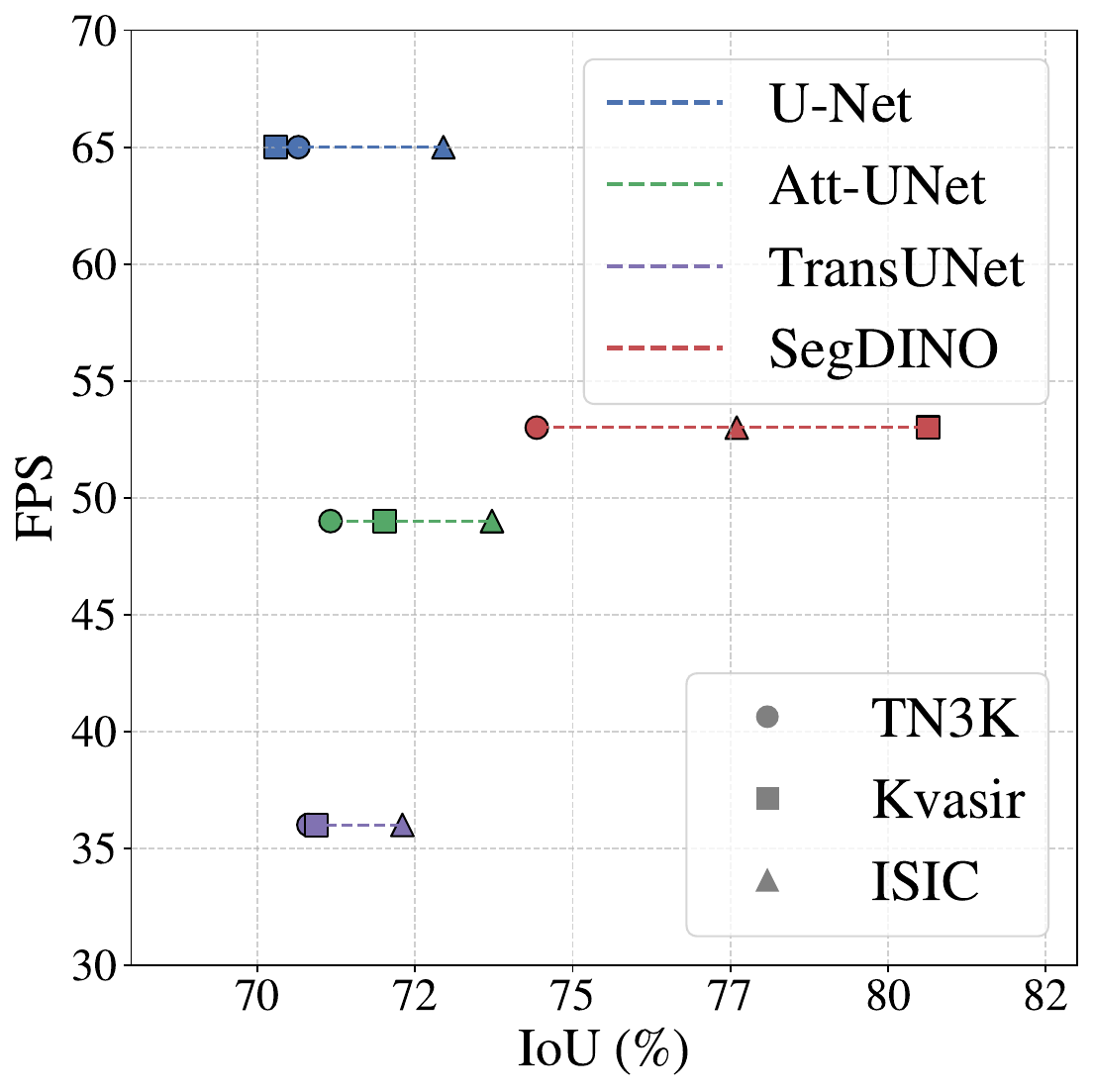}
    \label{fig:overview-fps}
\end{subfigure}
\caption{Overall performance and efficiency comparisons across different datasets.}
\label{fig:overview}
\end{figure}

\paragraph{Efficiency Comparisons.}
As illustrated in Fig.~\ref{fig:overview}, SegDINO demonstrates remarkable parameter efficiency while maintaining superior segmentation performance across both medical and natural datasets.
On Kvasir, SegDINO achieves the best performance with only 2.21M trainable parameters.
On the VMD-D dataset, SegDINO again delivers superior performance under a similarly compact parameter budget.
Moreover, SegDINO sustains an inference speed of 53 FPS, exceeding most transformer-based methods while being slightly lower than convolution-based architectures.
These results highlight that SegDINO consistently achieves the most favorable trade-off among performance, model size, and inference speed, establishing its advantage as a highly efficient solution for both medical and natural image segmentation.

\section{Conclusion}
In this work, we introduced SegDINO, a lightweight segmentation framework that couples a frozen DINOv3 backbone with a minimal MLP-based decoder. Our design directly addresses the long-standing challenge of adapting self-supervised representations to segmentation tasks without relying on heavy decoders. By reformulating multi-level patch tokens into a unified representation and employing an extremely light prediction head, SegDINO achieves strong segmentation accuracy while maintaining remarkable efficiency. 

Extensive experiments across six benchmarks, including three medical datasets (TN3K, Kvasir-SEG, ISIC) and three natural image datasets (MSD, VMD-D, ViSha), consistently demonstrate the advantages of our approach. SegDINO surpasses existing state-of-the-art models by large margins on both natural and medical image tasks, highlighting the effectiveness of leveraging foundation model features through a lightweight decoding pipeline. Notably, the results show that even a frozen DINOv3 backbone, when paired with a carefully designed light decoder, can outperform models that require significantly more parameters and computation. This validates our central hypothesis that decoder simplicity does not necessarily compromise segmentation performance when foundation features are properly exploited. 

Despite these strengths, SegDINO is not without limitations. First, by freezing the encoder, the adaptability of features to highly domain-specific distributions (\textit{e.g.}, rare pathological cases) may be constrained. Second, while our results confirm the benefits of lightweight decoding, further ablation studies are needed to better understand the contributions of individual components, such as feature selection depth, reformulation strategies, and decoder design. These analyses would provide deeper insights into the robustness of our framework and guide future architectural refinements.

\clearpage



\bibliography{iclr2026_conference}
\bibliographystyle{iclr2026_conference}
% \input{iclr2026_conference.bbl}

% \appendix
% \section{Appendix}

\end{document}